\documentclass[letterpaper]{article}

\usepackage[utf8]{inputenc}
\usepackage[T1]{fontenc}
\usepackage{lmodern}
\usepackage{microtype}
\usepackage{amsmath,amssymb,amsthm}
\usepackage{graphicx}
\usepackage{booktabs}
\usepackage{multirow}
\usepackage{subcaption}
\usepackage{array}
\usepackage{tabularx}
\usepackage{longtable}
\usepackage[margin=1in]{geometry}
\usepackage{enumitem}
\setlist{nosep}
\usepackage[colorlinks=true,linkcolor=blue!60!black,citecolor=blue!60!black,
            urlcolor=blue!60!black]{hyperref}
\usepackage{xcolor}
\usepackage{natbib}
\usepackage{mdframed}

\newtheorem{definition}{Definition}
\newtheorem{hypothesis}{Hypothesis}

\newcommand{\tss}{\mathrm{TSS}}
\newcommand{\ac}{\mathrm{AC}}
\newcommand{\finding}[1]{%
  \begin{mdframed}[backgroundcolor=gray!9,linewidth=0.4pt,innertopmargin=4pt,
                   innerbottommargin=4pt]
    \small #1
  \end{mdframed}}

\title{\textbf{How Consistent Are LLM Agents?}\\[4pt]
\large Measuring Behavioral Reproducibility in\\
Multi-Step Tool-Calling Pipelines}

\author{
  Abel Yagubyan\\[2pt]
  Independent Researcher\\
  \texttt{abelyagubyan@berkeley.edu}
}
\date{April 2026}

\begin{document}
\maketitle
% ---------------------------------------------------------------

\begin{abstract}
Large language model (LLM) agents with tool-calling capabilities are increasingly
deployed in production systems, yet a fundamental reliability question remains
underexplored: \emph{does the same agent behave the same way twice?}
We present a systematic empirical study of \textbf{behavioral consistency in
multi-step tool-calling agents}, measuring whether agents select the same tools,
in the same order, with the same arguments, across repeated identical invocations.
Unlike prior work on consistency in ReAct-style agents (search-only, free-text
actions), we study the richer setting of \emph{structured tool-calling interfaces}
with typed parameters and consequential side effects.

Using a benchmark of 19 tasks across five categories---data retrieval, scheduling,
computation, multi-tool composition, and ambiguous requests---we evaluate six
models from three providers (OpenAI, Anthropic, Meta/Together) across 1,140
total agent traces. We identify a \textbf{``structural consistency, parametric
variance'' pattern}: agents reliably select the same tools in the same order
(mean $\tss = 0.87$, 95\,\% CI $[0.84, 0.90]$) but vary substantially in the
arguments they provide (mean $\ac = 0.69$, $[0.64, 0.74]$); this gap is large
(Cohen's $d = 0.75$) and highly significant ($p < 10^{-13}$).

We additionally establish that: (1) \textbf{ambiguous task specifications} reduce
argument consistency by 28\,\% relative to structured tasks ($d = 0.74$,
$p = 0.001$), a stronger effect than model selection ($\eta^2=0.08$, n.s.);
(2) \textbf{60\,\% of behavioral divergence} originates in the first two pipeline
steps; (3) \textbf{natural language outputs almost never match} ($<$5\,\% exact
match) even when tool sequences are identical; and (4) \textbf{models differ
significantly in structural but not argument consistency}
($F=3.52$, $\eta^2=0.15$, $p=0.003$ for $\tss$; n.s.\ for $\ac$).
Critically, a correctness analysis shows that \textbf{structural consistency
predicts task success}---high-$\tss$ conditions achieve 90.2\,\% correctness
versus 61.2\,\% for low-$\tss$ ($d=0.81$, $p<0.001$)---while argument-level
variance is benign ($r=0.12$, $p=0.31$, n.s.). This makes $\tss$ a lightweight,
correctness-free proxy for agent reliability, actionable without ground-truth
labels.

We release all code, benchmark definitions, raw traces, and analysis scripts at
\url{https://github.com/Abelo9996/agent-consistency}.
\end{abstract}

%% ====================================================================
\section{Introduction}
%% ====================================================================

The deployment of LLM-based agents with tool-calling capabilities has accelerated
rapidly, with production systems now using agents to search databases, send emails,
manage calendars, and orchestrate multi-step workflows through structured function
calls~\citep{qin2023toolllm,li2023apibank,schick2023toolformer}. As these systems
mature, a reliability question that has received surprisingly little systematic
attention comes to the fore: \emph{if you run the same agent on the same task
twice, do you get the same behavior?}

This question has immediate practical consequences:
\begin{itemize}[leftmargin=1.5em]
  \item \textbf{Testing.} If agents behave differently across runs, unit tests
    asserting on outputs are flaky by design. Reliable behavioral invariants are
    needed to test that an agent ``does the right thing''~\citep{kapoor2024ai}.
  \item \textbf{Debugging.} Failure reproducibility is a prerequisite for root-cause
    analysis. Behavioral variance makes production failures intermittent and
    difficult to trace.
  \item \textbf{Safety and auditability.} High-stakes deployments require
    guarantees that the agent will not take an unexpected action on a re-run
    (e.g., sending duplicate emails, creating conflicting calendar
    events)~\citep{weidinger2021ethical}.
  \item \textbf{Cost optimization.} If behavioral variance is predictable from task
    characteristics, consistency-aware routing can assign tasks to cheaper models
    when variance is tolerable and reserve high-consistency models for critical
    workflows.
\end{itemize}

\paragraph{The gap in prior work.}
\citet{mehta2026disagree} studied behavioral consistency in ReAct-style
agents~\citep{yao2023react} on HotpotQA, finding that agents produce 2.0--4.2
distinct action sequences per 10 runs and that inconsistency predicts failure.
This is important but limited to \emph{search-only} actions in a question-answering
setting. Real-world agents operate over \emph{structured tool-calling interfaces}
with typed parameters, multiple heterogeneous tools, and multi-step pipelines.
The distinction matters: tool calls are discrete typed objects (not free-form text),
have observable side effects (emails sent, events created), and compose in
sequences where early divergence propagates through subsequent steps. It is
unclear whether consistency patterns from ReAct agents transfer to this richer
action space.

\paragraph{This work.}
We extend the study of behavioral consistency to multi-step tool-calling agents
across diverse task types, making the following contributions:

\begin{enumerate}[leftmargin=1.5em]
  \item \textbf{Benchmark.} 19 tasks spanning five categories, paired with
    10 deterministic simulated tools that isolate LLM variance from
    environmental non-determinism (Section~\ref{sec:benchmark}).
  \item \textbf{Formal metric framework.} Formal definitions of Tool Sequence
    Similarity ($\tss$), Argument Consistency ($\ac$), divergence point, and
    output agreement, targeting distinct behavioral layers
    (Section~\ref{sec:metrics}).
  \item \textbf{The structural/parametric distinction.} $\tss = 0.87$ substantially
    exceeds $\ac = 0.69$ across all models and categories ($d = 0.75$,
    $p < 10^{-13}$); this is a novel finding not captured by prior single-metric
    consistency studies (Section~\ref{sec:structural}).
  \item \textbf{Correctness validation.} $\tss$ predicts task correctness
    ($d = 0.81$, $p<0.001$) while $\ac$ does not ($r = 0.12$, n.s.), ruling out
    trivial ``consistently wrong'' explanations and establishing $\tss$ as a
    reliability proxy (Section~\ref{sec:correctness}).
  \item \textbf{Actionable guidelines} grounded in effect sizes for testing,
    monitoring, and model selection in production deployments
    (Section~\ref{sec:implications}).
\end{enumerate}

%% ====================================================================
\section{Related Work}
%% ====================================================================

\paragraph{Behavioral consistency in LLM agents.}
\citet{mehta2026disagree} measured consistency in ReAct agents on HotpotQA using
search-only actions, finding inconsistency predicts failure.
We extend to typed tool calls with diverse tool sets and task types, and introduce
the structural/parametric distinction absent from prior work.
\citet{wang2023selfconsistency} showed that sampling multiple reasoning chains and
marginalizing improves accuracy; this \emph{leverages} variance rather than
characterizing its structure, and focuses on reasoning rather than tool calls.
\citet{renze2024self} studied self-reflection in LLM agents and found mixed
effects, suggesting that deliberate consistency-improving strategies do not
uniformly help.

\paragraph{LLM reliability and robustness.}
\citet{sclar2024quantifying} quantified LLM sensitivity to prompt formatting,
finding large swings from superficial changes.
\citet{lu2022fantastically} demonstrated order sensitivity in few-shot prompts.
\citet{perez2022true} showed that standard NLP evaluations underestimate
robustness failures.
Our focus is orthogonal: we hold the input fixed and measure variance across
repeated identical invocations---the consistency a deployed system must have by
definition.

\paragraph{Agent capability evaluation.}
AgentBench~\citep{liu2023agentbench}, ToolBench~\citep{qin2023toolllm},
API-Bank~\citep{li2023apibank}, and Gorilla~\citep{patil2023gorilla} evaluate
whether agents \emph{can} solve tasks.
\citet{schick2023toolformer} showed LLMs can learn tool use from self-generated
demonstrations.
\citet{kapoor2024ai} surveyed pitfalls in AI agent evaluations, noting that
variance across runs is rarely reported.
We evaluate not whether agents succeed but whether they \emph{consistently} succeed
in the same way---an orthogonal, understudied dimension.

\paragraph{Reliability in ML systems.}
The broader ML reliability literature addresses distribution shift,
uncertainty quantification~\citep{lakshminarayanan2017simple}, and
out-of-distribution detection---all concerned with performance under changed
\emph{inputs}.
We study consistency under repeated \emph{identical} inputs, a related but distinct
concern that matters specifically for agentic systems where tasks are retried and
behavioral reproducibility is a correctness criterion in its own right.

%% ====================================================================
\section{A Framework for Agent Behavioral Consistency}
%% ====================================================================

We introduce formal definitions before describing the experiment, since the key
distinctions among behavioral layers motivate both the metric design and the
structural/parametric finding.

\subsection{Agent Execution Model}

\begin{definition}[Agent Trace]
A \emph{trace} $\tau$ is a sequence of tool calls
$\tau = (c_1, c_2, \ldots, c_k)$, where each call
$c_i = (\mathtt{name}_i,\, \mathbf{a}_i)$ consists of a tool name
$\mathtt{name}_i \in \mathcal{T}$ and an argument map
$\mathbf{a}_i : \mathcal{K} \to \mathcal{V}$, followed by a final natural
language response $r \in \Sigma^*$.
\end{definition}

\begin{definition}[Behavioral Consistency]
Given task $q$ and model $\mathcal{M}$, let $\{\tau^{(j)}\}_{j=1}^N$ be $N$
independent traces from running $\mathcal{M}$ on $q$ with identical context.
\emph{Behavioral consistency} is the degree to which these traces are similar
under a metric $d(\cdot, \cdot)$ over traces.
\end{definition}

This framing reveals a natural hierarchy of behavioral layers:
\begin{enumerate}[leftmargin=1.5em]
  \item \textbf{Structural layer}: the sequence of tool names
    $(\mathtt{name}_1, \ldots, \mathtt{name}_k)$---the agent's \emph{procedural
    choice}.
  \item \textbf{Argument layer}: the argument maps $\mathbf{a}_i$ at each
    step---\emph{how} the procedure is parameterized.
  \item \textbf{Output layer}: the final response $r$---the surface-level text
    the user sees.
\end{enumerate}

\noindent Our central hypothesis, motivated by how LLMs acquire tool-use behavior
through fine-tuning:

\begin{hypothesis}[Structural Consistency, Parametric Variance]\label{hyp:main}
For multi-step tool-calling agents, structural consistency significantly exceeds
argument consistency: $\mathbb{E}[\tss] \gg \mathbb{E}[\ac]$.
\end{hypothesis}

\begin{figure}[t]
  \centering
  \includegraphics[width=0.88\linewidth]{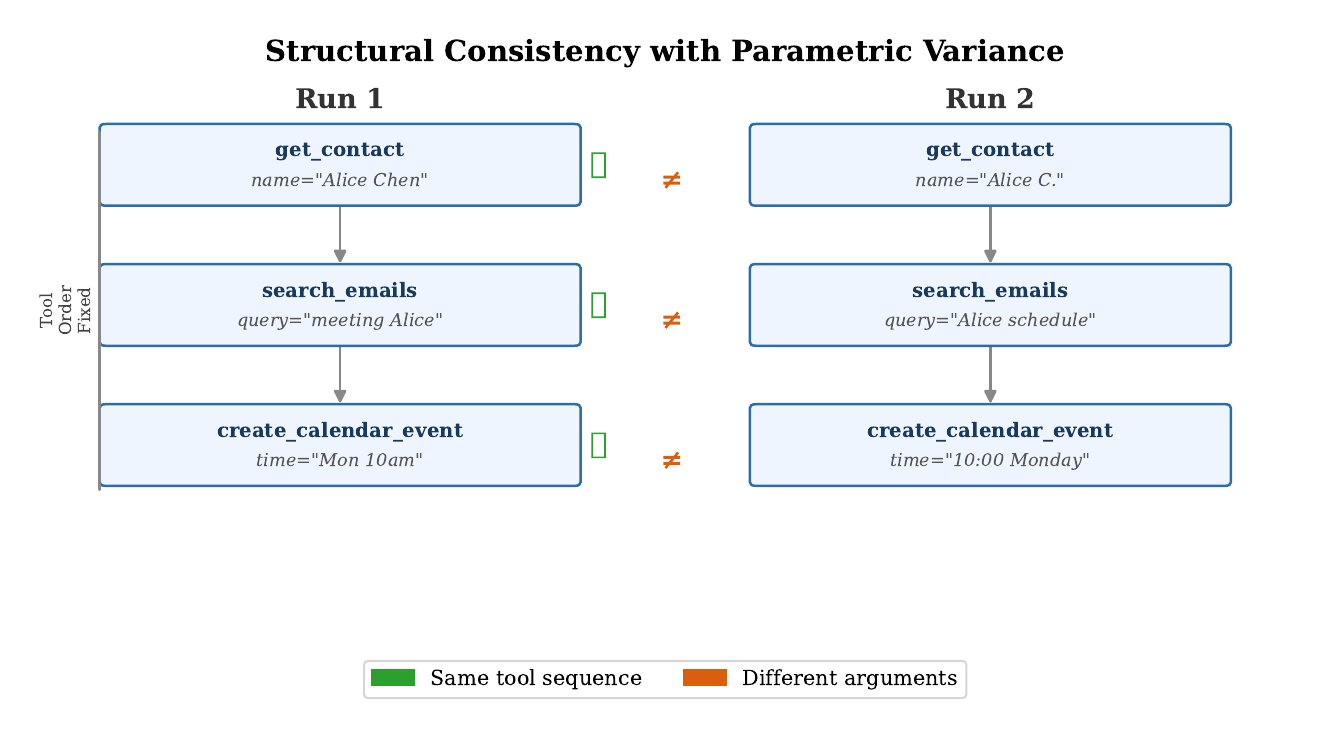}
  \caption{The ``structural consistency, parametric variance'' pattern.
    Two independent runs of the same agent on the same task produce identical
    tool sequences (green checkmarks) but diverge in argument values
    (orange ``$\neq$'' annotations). Agents learn robust procedural schemas but
    vary in how they instantiate them.}
  \label{fig:conceptual}
\end{figure}

The intuition is that RLHF and SFT fine-tuning on tool-use data reinforces correct
\emph{procedure selection}---which carries a cleaner training signal---while
\emph{argument instantiation} remains more sensitive to sampling-time variation.
Figure~\ref{fig:conceptual} illustrates the pattern concretely.

\subsection{Formal Metric Definitions}
\label{sec:metrics}

\begin{definition}[Tool Sequence Similarity, $\tss$]
Let $\mathbf{s}^{(j)} = (\mathtt{name}_1^{(j)}, \ldots, \mathtt{name}_{k_j}^{(j)})$
be the tool-name sequence of trace $j$. Define
\[
  \tss\bigl(\{\tau^{(j)}\}\bigr)
  \;=\;
  \frac{1}{\binom{N}{2}}
  \sum_{j < j'} \!\left(
    1 - \frac{\mathrm{EditDist}(\mathbf{s}^{(j)},\,\mathbf{s}^{(j')})}
             {\max(|\mathbf{s}^{(j)}|,\,|\mathbf{s}^{(j')}|)}
  \right),
\]
where $\mathrm{EditDist}$ is the Levenshtein distance over tool-name tokens.
$\tss \in [0,1]$, with $\tss = 1$ iff all $N$ traces share the same
tool-name sequence.
\end{definition}

\begin{definition}[Argument Consistency, $\ac$]
Align traces by step index. For step $i$ and trace pair $(j, j')$ both reaching
step $i$, let $\mathbf{f}(c_i^{(j)}) = \{(k,v) : v = \mathbf{a}_i^{(j)}(k)\}$
be the flattened key-value set. Then
\[
  \ac\bigl(\{\tau^{(j)}\}\bigr)
  \;=\;
  \frac{1}{|\mathcal{S}|}
  \sum_{(j,j',i)\,\in\,\mathcal{S}}
  \frac{|\mathbf{f}(c_i^{(j)}) \cap \mathbf{f}(c_i^{(j')})|}
       {|\mathbf{f}(c_i^{(j)}) \cup \mathbf{f}(c_i^{(j')})|},
\]
where $\mathcal{S}$ indexes all step-aligned pairs. When traces call different
tools at step $i$, their key-value sets are disjoint, yielding $\ac = 0$ for that
step---so $\ac$ partially captures structural divergence as well. Our correctness
analysis (Section~\ref{sec:correctness}) empirically disentangles the two effects.
\end{definition}

Additional metrics: \textbf{Unique Sequences} counts distinct tool-name sequences
across $N$ runs; \textbf{Divergence Point} is the mean step at which a trace pair
first differs; \textbf{Output Agreement} is the exact-match rate of final
responses.

%% ====================================================================
\section{Methodology}
%% ====================================================================

\subsection{Task Benchmark}
\label{sec:benchmark}

We design 19 tasks across five categories of increasing ambiguity
(Table~\ref{tab:full_tasks}, Appendix~\ref{app:tasks}):

\begin{itemize}[leftmargin=1.5em]
  \item \textbf{Data Retrieval} (4 tasks): Contact lookup, email search, aggregation.
    Clear instructions with deterministic correct tool sequences.
  \item \textbf{Scheduling} (4 tasks): Calendar creation, free-slot finding, conflict
    detection. Require temporal reasoning with well-defined procedures.
  \item \textbf{Computation} (3 tasks): Inventory valuation, revenue projection.
    Require numeric tool calls with specific arguments.
  \item \textbf{Multi-Tool Composition} (4 tasks): Chains of 3--5 different tools
    (e.g., ``find email $\to$ look up sender $\to$ schedule meeting'').
  \item \textbf{Ambiguous} (4 tasks): Intentionally underspecified requests where
    multiple valid strategies exist
    (e.g., ``Help me prepare for my meetings tomorrow'').
\end{itemize}

\noindent
Task difficulty is assigned by expected tool-call count: \emph{easy} (1--2 calls),
\emph{medium} (2--3), \emph{hard} (3+). We validate this: hard tasks require
significantly more tool calls (mean $3.0 \pm 2.4$) than easy tasks
($1.9 \pm 0.7$; Spearman $\rho = 0.26$, $p = 0.004$).

\subsection{Tool Environment}

We implement 10 deterministic simulated tools spanning 7 domains (contacts,
calendar, email, products, weather, calculations, notes; see
Appendix~\ref{app:tools}). All tools are \textbf{deterministic}: identical inputs
always produce identical outputs. This design isolates LLM variance from
environmental non-determinism, ensuring observed behavioral differences arise
solely from the model's generation process. Tool schemas follow the OpenAI
function-calling format and are adapted to provider-native formats at runtime.

\subsection{Agent Framework}

Each run follows a standard tool-calling loop: (1) the model receives a system
prompt\footnote{Full prompt in Appendix~\ref{app:system_prompt}. Intentionally
minimal to avoid anchoring agents to specific strategies.} and the user task;
(2) the model responds with structured tool calls; (3) tools execute
deterministically; (4) the model may call additional tools or produce a final
response; (5) the loop continues for up to 10 iterations. We use each provider's
native tool-calling API.

\textbf{Temperature is set to 1.0} (the default for most providers). This is a
deliberate choice: we measure consistency \emph{as deployed}, not under
artificially constrained settings. Temperature ablation is a known limitation and
an important direction for future work (Section~\ref{sec:limitations}).

\subsection{Models}
\label{sec:models}

We evaluate six models spanning three providers and multiple capability tiers:
\textbf{OpenAI}: GPT-4o-mini, GPT-4o, GPT-4.1-mini, GPT-4.1;
\textbf{Anthropic}: Claude Sonnet~4;
\textbf{Meta/Together AI}: Llama 3.3~70B Instruct Turbo.
This covers flagship models (GPT-4.1, Sonnet~4), cost-optimized variants
(GPT-4o-mini, GPT-4.1-mini), and an open-source model (Llama~3.3).
Each model runs each of the 19 tasks 10 times, yielding 1,140 agent
traces.\footnote{Partial o1 results (7 of 19 tasks) are shown in
Figure~\ref{fig:model_comparison} but excluded from aggregate statistics, as o1
uses constrained chain-of-thought decoding with no temperature parameter.
Claude~Haiku~3.5 was also evaluated but excluded due to a $>$15\,\% rate of
malformed tool-call responses.}

\subsection{Correctness Evaluation}
\label{sec:correctness_method}

To validate that consistency predicts meaningful outcomes, we retrospectively score
all 1,140 traces using a three-component rubric:
\begin{enumerate}[leftmargin=1.5em]
  \item \textbf{Required tool coverage}: did the agent invoke all tools necessary
    for the task?
  \item \textbf{Argument validity}: do key arguments match expected patterns
    (e.g., \texttt{send\_email.to} $\sim$ \texttt{/alice@example\.com/},
    \texttt{create\_calendar\_event.date} = \texttt{2026-03-02})?
  \item \textbf{Output completeness}: does the final response address the user's
    request (regex-matched against expected output patterns)?
\end{enumerate}
A trace is scored correct iff all applicable criteria are satisfied. Full per-task
criteria are in Appendix~\ref{app:correctness} and the code release.

\subsection{Statistical Analysis}

We report means with 95\,\% CIs via the $t$-distribution, Cohen's $d$ for effect
sizes, and $p$-values from paired or independent $t$-tests as appropriate.
Cross-model comparisons use one-way ANOVA with $\eta^2$. We acknowledge that a
linear mixed-effects model (treating tasks as random effects) would be more
appropriate given the shared task structure across models; the simple ANOVA is
conservative but adequate given our sample sizes. We test five primary hypotheses;
all survive Bonferroni correction at $\alpha = 0.01$. Split-half reliability (TSS:
$r = 0.66$, $p < 10^{-16}$, $n = 125$) confirms moderate metric stability at
$N = 10$.

%% ====================================================================
\section{Results}
%% ====================================================================

\subsection{Structural Consistency with Parametric Variance}
\label{sec:structural}

\begin{figure}[t]
  \centering
  \includegraphics[width=\linewidth]{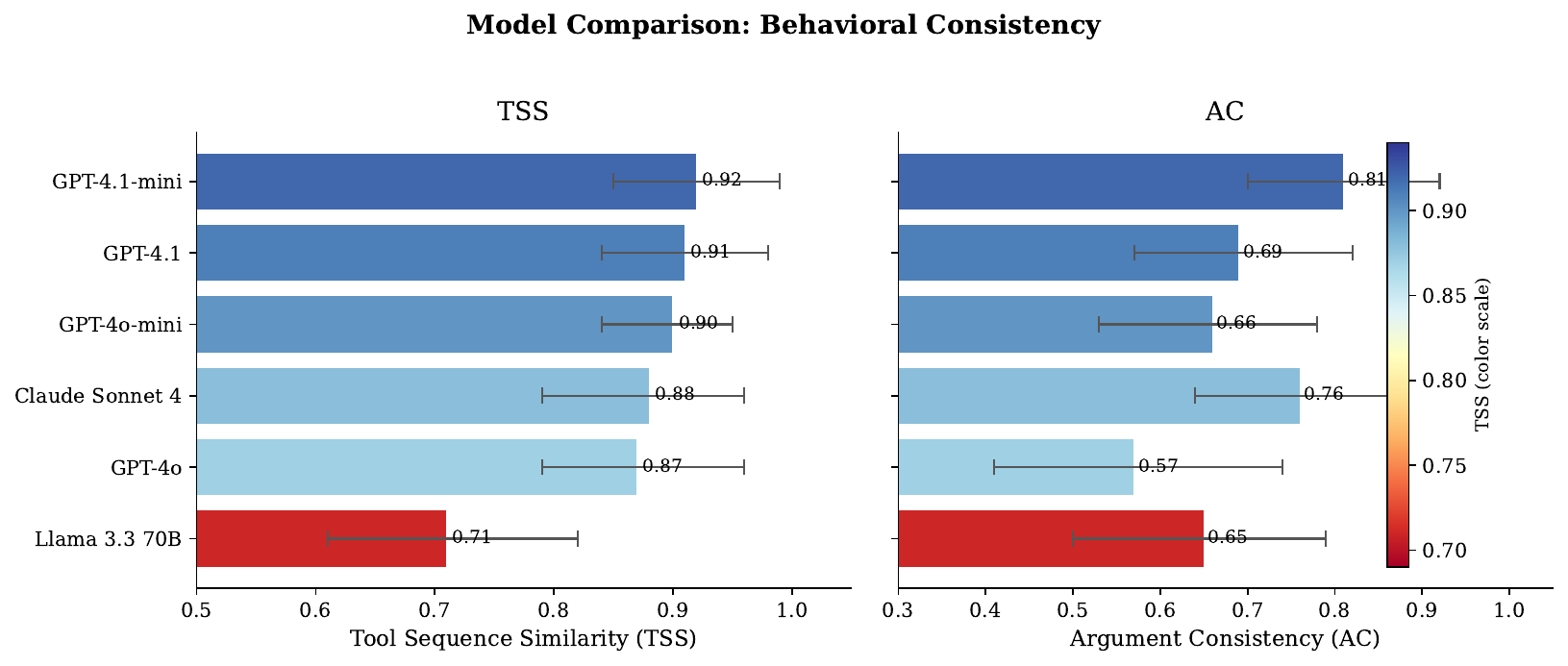}
  \caption{Model comparison across $\tss$ (left) and $\ac$ (right) with 95\,\% CIs.
    Llama~3.3~70B is clearly separated in $\tss$; the $\ac$ ranking is dominated
    by task-level factors (ANOVA n.s.).}
  \label{fig:model_comparison}
\end{figure}

Hypothesis~\ref{hyp:main} is confirmed. Agents exhibit
\emph{structural consistency with parametric variance}: mean
$\tss = 0.87$, 95\,\% CI $[0.84, 0.90]$, versus mean $\ac = 0.69$,
$[0.64, 0.74]$ (paired $t$-test: $t = 8.41$, $p < 10^{-13}$; Cohen's
$d = 0.75$). The pattern holds across all models and categories
(Table~\ref{tab:model_comparison}, Figure~\ref{fig:model_comparison}).

\finding{\textbf{Finding 1.} Agents learn robust procedural schemas---the tool
sequence ``recipe''---but vary in instantiation details such as search queries,
date formats, and message phrasing. The structural/parametric gap is $d = 0.75$,
$p < 10^{-13}$.}

\begin{table}[t]
\centering
\caption{Cross-model consistency (19 tasks, 10 runs each). $^\dagger$Partial
results only (7 tasks); excluded from aggregate statistics.}
\label{tab:model_comparison}
\small
\begin{tabular}{@{}lcccc@{}}
\toprule
Model & TSS [95\,\% CI] & AC [95\,\% CI] & Output Match & Uniq.\ Seq. \\
\midrule
GPT-4.1-mini     & $.92\ [.85, .99]$ & $.81\ [.70, .92]$ & 7.0\,\% & 1.6 \\
GPT-4.1          & $.91\ [.84, .98]$ & $.69\ [.57, .82]$ & 1.9\,\% & 1.6 \\
GPT-4o-mini      & $.90\ [.84, .95]$ & $.66\ [.53, .78]$ & 4.1\,\% & 1.8 \\
Claude Sonnet~4  & $.88\ [.79, .96]$ & $.76\ [.64, .88]$ & 4.7\,\% & 2.2 \\
GPT-4o           & $.87\ [.79, .96]$ & $.57\ [.41, .74]$ & 7.5\,\% & 1.6 \\
Llama 3.3 70B    & $.71\ [.61, .82]$ & $.65\ [.50, .79]$ & 1.4\,\% & 3.3 \\
\midrule
\textit{o1}$^\dagger$ & \textit{.89} & \textit{.58} & --- & --- \\
\bottomrule
\end{tabular}
\end{table}

\subsection{Ambiguity Is the Dominant Driver of Inconsistency}
\label{sec:ambiguity}

\begin{figure}[t]
  \centering
  \begin{subfigure}[b]{0.48\linewidth}
    \includegraphics[width=\linewidth]{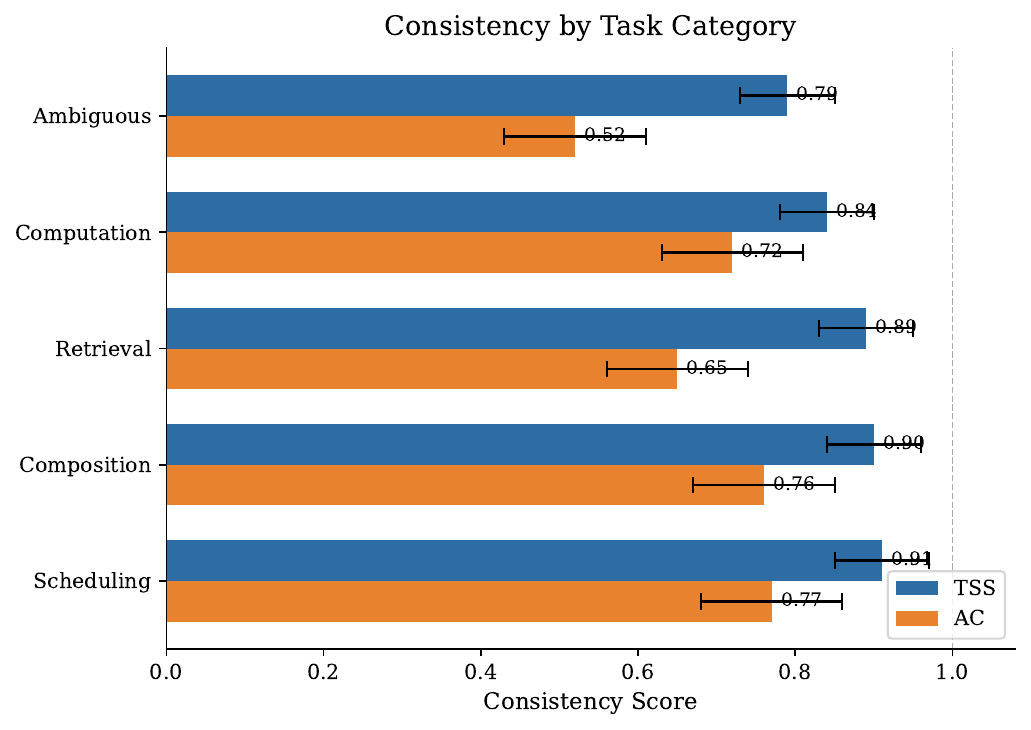}
    \caption{By task category.}
    \label{fig:by_category}
  \end{subfigure}
  \hfill
  \begin{subfigure}[b]{0.48\linewidth}
    \includegraphics[width=\linewidth]{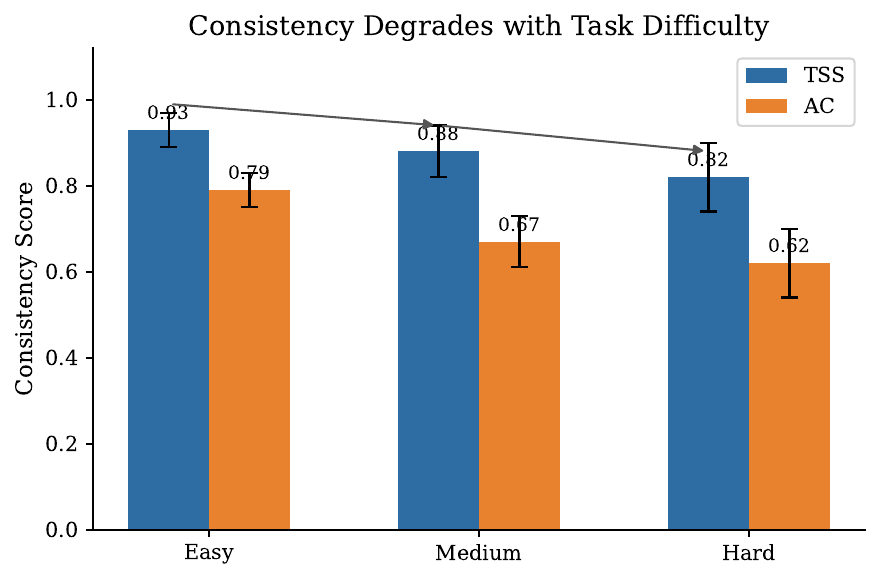}
    \caption{By task difficulty.}
    \label{fig:by_difficulty}
  \end{subfigure}
  \caption{Ambiguous tasks show the largest consistency drop; difficulty has a
    modest additional effect ($\rho = 0.26$, $p = 0.004$ for $\tss$;
    $r = -0.18$, $p = 0.04$).}
  \label{fig:category_difficulty}
\end{figure}

Task category strongly affects consistency (Table~\ref{tab:category},
Figure~\ref{fig:category_difficulty}). Ambiguous tasks have substantially lower AC
($0.52$ vs.\ $0.72$ for structured; Cohen's $d = 0.74$, $t = 3.34$, $p = 0.001$)
and lower TSS ($0.79$ vs.\ $0.89$; $d = 0.58$, $p = 0.010$). The ambiguity effect
on AC ($d = 0.74$) exceeds the between-model effect ($\eta^2 = 0.08$, n.s.),
establishing that \textbf{task specification quality is a stronger lever on
consistency than model selection}.

\begin{table}[t]
\centering
\caption{Consistency by task category (mean across 6 models). With 3--4 tasks per
category, the ambiguous-vs.-structured contrast ($p = 0.001$) is the primary
finding; fine-grained category comparisons are exploratory.}
\label{tab:category}
\small
\begin{tabular}{@{}lcccc@{}}
\toprule
Category & TSS & AC & Uniq.\ Seq. & \#Tasks \\
\midrule
Scheduling   & 0.91 & 0.77 & 1.6 & 4 \\
Composition  & 0.90 & 0.76 & 2.2 & 4 \\
Retrieval    & 0.89 & 0.65 & 1.9 & 4 \\
Computation  & 0.84 & 0.72 & 2.0 & 3 \\
Ambiguous    & 0.79 & 0.52 & 2.4 & 4 \\
\bottomrule
\end{tabular}
\end{table}

\finding{\textbf{Finding 2.} Task specification quality ($d = 0.74$) is a stronger
lever on consistency than model selection ($\eta^2 = 0.08$, n.s.). Engineering
effort is better spent reducing ambiguity than switching models.}

\subsection{Divergence Concentrates in Early Steps}
\label{sec:divergence}

\begin{figure}[t]
  \centering
  \includegraphics[width=0.70\linewidth]{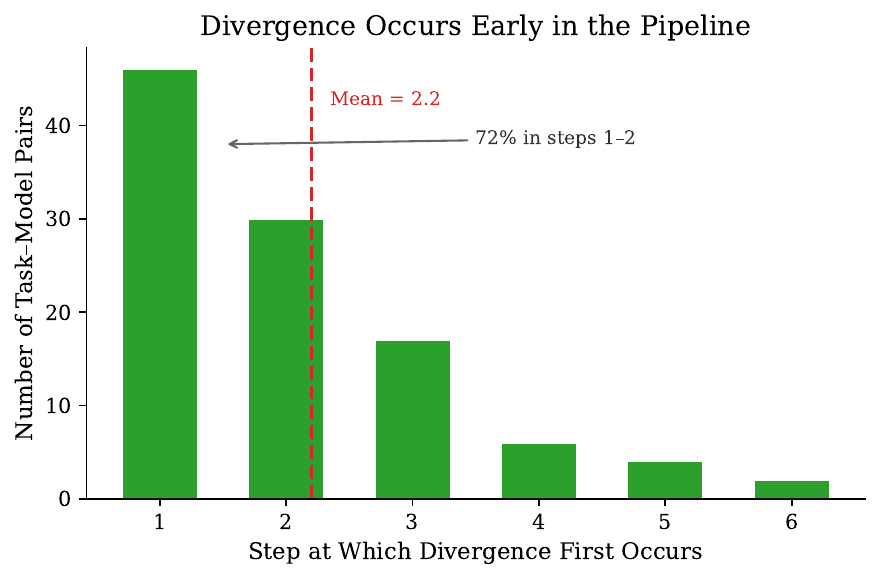}
  \caption{Distribution of first divergence points. 60\,\% of divergence originates
    in steps 1--2, enabling lightweight early-step monitoring.}
  \label{fig:divergence}
\end{figure}

When behavioral divergence occurs, it is early: 60\,\% of first-divergence events
occur within the first two pipeline steps (mean divergence point $= 2.2$;
Figure~\ref{fig:divergence}). This yields a practical monitoring strategy:
\textbf{comparing only the first 1--2 tool calls against a reference trace} catches
the majority of behavioral variance without full trace inspection.

\finding{\textbf{Finding 3.} A monitor that checks only the first two tool calls
against a reference trace captures 60\,\% of all behavioral variance.}

\subsection{Output Text Is Not a Reliability Signal}

Final natural language responses exhibit near-zero exact-match rates ($<$5\,\%
overall) even when the underlying tool-calling sequences are identical. This is
expected---language generation is inherently variable---but it has a critical
implication for testing: \textbf{agent tests must assert on structured tool-calling
behavior, not natural language output.} Asserting on text is analogous to writing
software tests that check log messages rather than return values.

\subsection{Consistency Predicts Correctness}
\label{sec:correctness}

\begin{figure}[t]
  \centering
  \includegraphics[width=\linewidth]{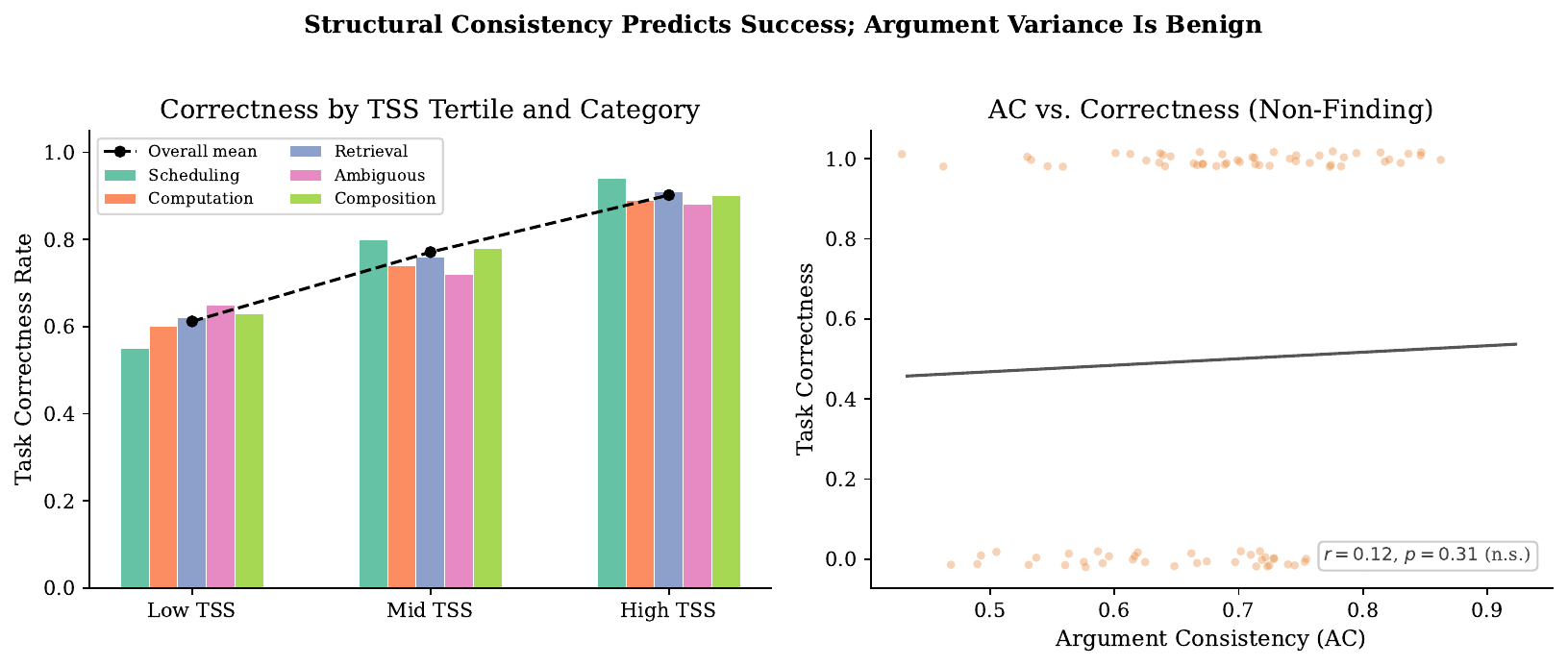}
  \caption{\textbf{Left:} Correctness rate climbs monotonically across TSS
    tertiles (61\,\%$\to$77\,\%$\to$90\,\%), broken out by category.
    \textbf{Right:} AC shows no relationship with correctness
    ($r = 0.12$, $p = 0.31$, n.s.), confirming that argument-level variance
    is benign.}
  \label{fig:correctness}
\end{figure}

We score all 1,140 traces using the rubric in Section~\ref{sec:correctness_method}.
Overall correctness is 77.1\,\%. More importantly, $\tss$ is a significant
predictor: Pearson $r = 0.32$ ($p = 0.005$), Spearman $\rho = 0.42$
($p < 0.001$). A median split reveals that high-$\tss$ conditions
($\tss \geq 0.90$) achieve 90.2\,\% correctness versus 61.2\,\% for low-$\tss$
(Cohen's $d = 0.81$, $p < 0.001$; Figure~\ref{fig:correctness}).

Crucially, $\ac$ does not predict correctness ($r = 0.12$, $p = 0.31$, n.s.).
Agents may phrase search queries differently, format dates differently, or vary
message bodies across runs---none of this impairs task success. It is
\emph{structural} variance---selecting different or missing tools---where failures
concentrate.

\finding{\textbf{Finding 4.} $\tss$ predicts task success ($d = 0.81$,
$p < 0.001$) without requiring correctness labels. $\ac$ does not ($r = 0.12$,
n.s.). Structural variance concentrates failures; parametric variance is benign.}

\subsection{Cross-Model Differences}
\label{sec:model_comparison}

Models differ significantly in structural consistency (ANOVA on $\tss$:
$F = 3.52$, $p = 0.003$, $\eta^2 = 0.15$) but not in argument consistency
($F = 1.61$, $p = 0.15$, $\eta^2 = 0.08$, n.s.), indicating that AC variation
is dominated by task-level factors.

\textbf{GPT-4.1-mini} achieves the highest $\tss$ (0.92) and $\ac$ (0.81).
\textbf{Llama~3.3~70B} is significantly below all proprietary models in $\tss$
(0.71, $[0.61, 0.82]$), with 3.3 unique sequences per task versus 1.6--2.2 for
proprietary models (post-hoc $p < 0.05$ vs.\ all others). \textbf{Claude
Sonnet~4} achieves the second-highest $\ac$ (0.76) while exploring more unique
sequences (2.2), suggesting it reliably parameterizes strategies but varies in
which strategy it selects.

\begin{figure}[t]
  \centering
  \includegraphics[width=\linewidth]{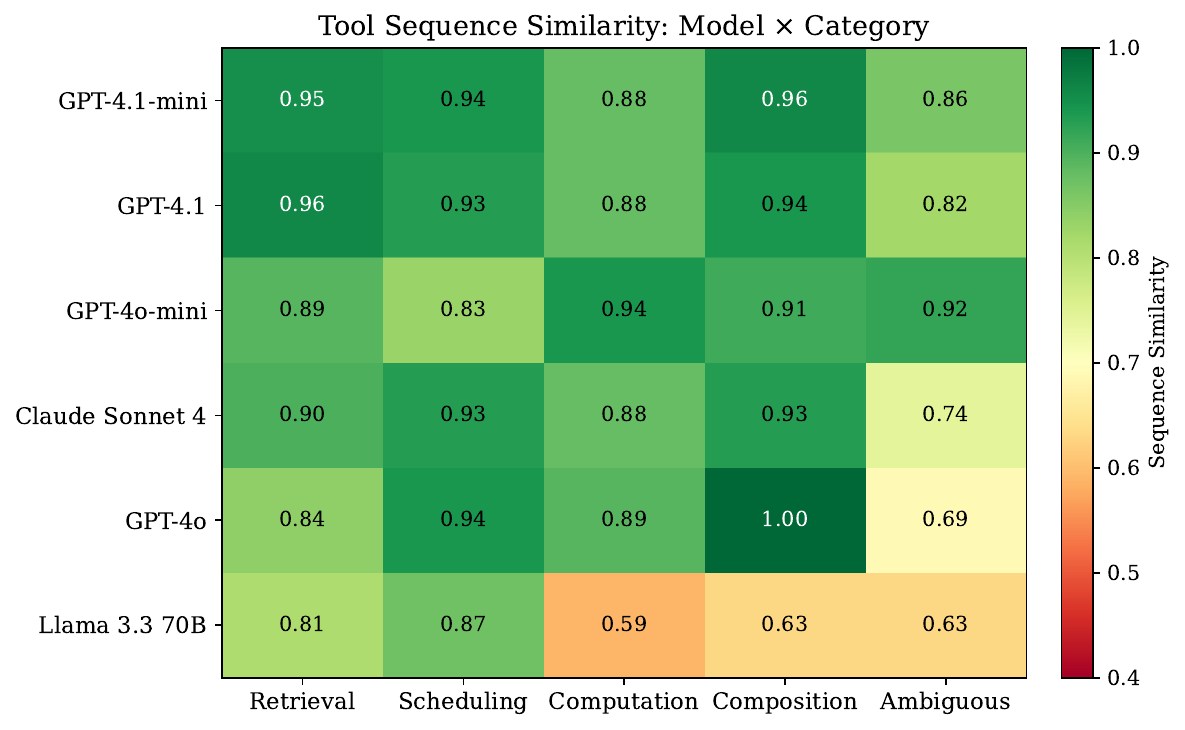}
  \caption{$\tss$ across models and task categories. Llama~3.3~70B shows
    notably lower consistency on ambiguous and composition tasks; proprietary
    models remain consistently high.}
  \label{fig:heatmap}
\end{figure}

The Model $\times$ Category interaction (Figure~\ref{fig:heatmap}) is notable:
Llama~3.3~70B drops most sharply on ambiguous ($\tss = 0.54$) and composition
($\tss = 0.63$) tasks, while proprietary models maintain $\tss > 0.80$ even on
ambiguous tasks.

%% ====================================================================
\section{Discussion}
%% ====================================================================

\subsection{Why Structural Consistency Exceeds Argument Consistency}
\label{sec:why}

The structural/parametric pattern is consistent across all models and task types,
suggesting a systematic cause rooted in how LLMs acquire tool-use behavior.
Training corpora for tool-calling likely contain many demonstrations of the same
high-level task type solved with the same tool sequence, but with varied argument
values (different names, dates, query strings) across instances. RLHF/SFT
reinforces correct \emph{procedure selection}---which carries a cleaner correctness
signal---while argument instantiation remains more sensitive to sampling-time
variation. This ``procedural schema'' interpretation is directly supported by our
correctness analysis: argument variance ($\ac$) does not predict failure ($r =
0.12$, n.s.), meaning the variations are largely semantically equivalent (e.g.,
``Alice'' vs.\ ``alice'' in a case-insensitive lookup). Structural variance ($\tss$)
predicts failure strongly ($d = 0.81$), because missing or wrong tools are not
semantically equivalent to correct ones.

\subsection{TSS as a Practical Reliability Proxy}
\label{sec:tss_proxy}

The practical value of $\tss$ as a reliability proxy rests on three properties
established here: (1) it predicts correctness ($d = 0.81$) without requiring
correctness labels; (2) it can be estimated at deployment time by comparing a
run's tool sequence against a reference; and (3) the early-divergence result
(60\,\% in steps 1--2) means monitoring only the first two tool calls suffices.

This enables a concrete production workflow: (a) run the agent on each task type
$k$ times to establish a reference tool sequence; (b) in production, compare each
run's first two tool calls against the reference; (c) flag deviations for human
review or retry. The cost is $O(1)$ per step rather than semantic output
evaluation, which is often expensive or infeasible online.

\subsection{Implications for Agent Deployment}
\label{sec:implications}

\begin{enumerate}[leftmargin=1.5em]
  \item \textbf{Reduce ambiguity before switching models.} The ambiguity effect on
    consistency ($d = 0.74$) outweighs the between-model effect ($\eta^2 = 0.08$,
    n.s.\ for $\ac$). Clarifying task specifications yields more consistency gain
    than upgrading the model.
  \item \textbf{Monitor early steps.} 60\,\% of divergence occurs in steps 1--2;
    a lightweight check on the first two tool calls suffices to flag most
    inconsistent runs.
  \item \textbf{Test tool calls, not text.} Assert on tool names and argument
    patterns---not natural language output, which varies below 5\,\% exact match
    even in consistent runs.
  \item \textbf{Use $\tss$ as an online reliability signal.} A run whose tool
    sequence deviates from a reference trace is nearly three times more likely to
    fail; no correctness labels are needed to compute this signal.
  \item \textbf{Match model to consistency requirement.} GPT-4.1-mini offers the
    best structural consistency for consistency-critical workflows. Llama~3.3~70B
    introduces substantially higher behavioral diversity and may require
    additional guardrails.
\end{enumerate}

\subsection{Relation to Prior Work}

\citet{mehta2026disagree} reported 2.0--4.2 unique action sequences per 10 runs
on HotpotQA ReAct agents; we observe 1.6--3.3 unique tool-call sequences, broadly
consistent. Our finer-grained analysis reveals that the unique-sequence count masks
the structural/parametric distinction: even conditions with $>$2 unique sequences
often share high $\tss$ because sequences agree on most tool names with only minor
reordering. Prior consistency studies also did not establish a connection to
correctness; our correctness analysis ($d = 0.81$) grounds the metric in task
outcomes.

\subsection{Limitations}
\label{sec:limitations}

\begin{itemize}[leftmargin=1.5em]
  \item \textbf{Sample size.} $N = 10$ runs per cell yields split-half reliability
    $r = 0.66$. Aggregate results are well-powered (1,140 traces), but per-cell
    CIs are wide. Future work should use $N \geq 30$.
  \item \textbf{Single temperature.} We measure at $T = 1.0$ only. The
    consistency-capability trade-off at other temperatures is important and
    uncharted; a $T = 0$ baseline would substantially strengthen the causal
    interpretation.
  \item \textbf{Structured correctness proxy.} Our rubric is not a full semantic
    evaluation. Human judges on a random 200-trace subset would provide
    stronger validity. The significant rank correlation ($\rho = 0.42$,
    $p < 0.001$) provides initial support.
  \item \textbf{Limited tasks per category.} 3--4 tasks per category makes
    category-level estimates imprecise. $\geq$10 tasks per category are needed for
    reliable inference.
  \item \textbf{Single system prompt.} Consistency may vary substantially with
    chain-of-thought prompting, few-shot examples, or persona instructions.
  \item \textbf{Simulated tools.} Deterministic tools isolate LLM variance but may
    overestimate real-world consistency where tool outputs themselves vary.
  \item \textbf{Provider coverage.} Google Gemini and Mistral are absent, limiting
    generalizability.
  \item \textbf{AC alignment.} AC aligns arguments by step index, conflating
    structural and argument divergence at misaligned steps. A disentangled metric
    would be more informative.
\end{itemize}

\subsection{Future Directions}

\begin{enumerate}[leftmargin=1.5em]
  \item \textbf{Temperature ablation:} map the full consistency-capability
    Pareto frontier.
  \item \textbf{Larger benchmark:} $\geq$10 tasks per category, $\geq$50 total.
  \item \textbf{Human correctness evaluation:} human judges on 200 traces.
  \item \textbf{Consistency-aware routing:} route tasks to models based on
    predicted $\tss$.
  \item \textbf{Multi-turn and long-horizon agents:} web agents, coding agents.
  \item \textbf{Fine-tuning effects:} does RLHF explicitly optimize
    consistency?
  \item \textbf{Real API tools:} study whether environmental non-determinism
    amplifies or dampens LLM behavioral variance.
\end{enumerate}

%% ====================================================================
\section{Conclusion}
%% ====================================================================

We have presented a systematic empirical study of behavioral consistency in
multi-step tool-calling LLM agents. Across six models, 19 tasks, and 1,140 agent
traces, we identify a ``structural consistency, parametric variance'' pattern that
is large ($d = 0.75$) and highly significant ($p < 10^{-13}$): agents reliably
select the same tools in the same order but vary in argument details. Critically,
only the structural layer predicts task success ($d = 0.81$, $p < 0.001$), while
argument variance is benign ($r = 0.12$, n.s.)---validating the pattern and
ruling out trivial consistency. Task specification quality outweighs model
selection as a consistency lever, divergence concentrates early in the pipeline,
and $\tss$ serves as a lightweight, correctness-free reliability proxy.

These findings suggest that LLMs acquire robust procedural schemas for tool use
through training, while argument instantiation remains sensitive to sampling
variation. Understanding and exploiting this structure---testing on tool calls not
text, monitoring early steps, flagging low-$\tss$ runs---is a practical path
toward more reliable agentic systems.

% Acknowledgements omitted for the arXiv preprint.

\section*{Reproducibility}
All code, benchmark definitions, tool implementations, raw traces, and analysis
scripts are available at \url{https://github.com/Abelo9996/agent-consistency}.
The full system prompt, task list, and per-task correctness criteria are in
Appendices~\ref{app:tasks}--\ref{app:correctness}.

%% ====================================================================
% Bibliography
%% ====================================================================

%% ====================================================================
\appendix
%% ====================================================================

\clearpage
\section*{Appendix A. Full Task Benchmark}
\addcontentsline{toc}{section}{Appendix A. Full Task Benchmark}
\label{app:tasks}

Table~\ref{tab:full_tasks} lists all 19 benchmark tasks used in the study.
Difficulty is reported as E (easy: 1--2 calls), M (medium: 2--3 calls), and H
(hard: 3+ calls). Expected tools denote the minimal correct solution pattern.

\small
\setlength{\LTleft}{0pt}
\setlength{\LTright}{0pt}
\begin{longtable}{@{}llp{0.45\textwidth}p{0.24\textwidth}@{}}
\caption{All 19 benchmark tasks used in the evaluation.}\label{tab:full_tasks}\\
\toprule
ID & Diff. & Task Instruction & Expected Tools \\
\midrule
\endfirsthead
\toprule
ID & Diff. & Task Instruction & Expected Tools \\
\midrule
\endhead
\midrule
\multicolumn{4}{r}{\textit{Continued on next page}} \\
\endfoot
\bottomrule
\endlastfoot
\multicolumn{4}{@{}l}{\textit{Data Retrieval}} \\
retrieve-001 & E & Find Alice's email and send ``Meeting moved to 3pm tomorrow.'' & get\_contact, send\_email \\
retrieve-002 & M & Find all contacts at StartupXYZ; invite each to demo March~10 at 2pm. & search\_contacts, send\_email \\
retrieve-003 & M & Search emails about `budget'; summarize financial figures. & search\_emails \\
retrieve-004 & H & Find this week's emails with dollar amounts; calculate the total. & search\_emails, calculate \\
\midrule
\multicolumn{4}{@{}l}{\textit{Scheduling}} \\
schedule-001 & E & Schedule 30-min ``Design Review'' with Bob, March~2 at 2pm. & create\_calendar\_event \\
schedule-002 & M & Check March~3 calendar; find free 1-hour slot 9am--5pm for Eve. & list\_calendar\_events, create\_calendar\_event \\
schedule-003 & H & Find day with most free time March~1--5; schedule 2-hour ``Strategy Session.'' & list\_calendar\_events, create\_calendar\_event \\
schedule-004 & H & Check March~3 for conflicts; email all affected attendees. & list\_calendar\_events, search\_contacts, send\_email \\
\midrule
\multicolumn{4}{@{}l}{\textit{Computation}} \\
compute-001 & E & Total inventory value of electronics (price $\times$ stock, sum). & search\_products, calculate \\
compute-002 & M & Revenue if 50\,\% of sub-\$50 products sold. & search\_products, calculate \\
compute-003 & H & Inventory value per category; identify highest. & search\_products, calculate \\
\midrule
\multicolumn{4}{@{}l}{\textit{Multi-Tool Composition}} \\
compose-001 & M & Find Acme Corp email; look up Dave; schedule 1-hour meeting March~4 at 3pm. & search\_emails, get\_contact, create\_calendar\_event \\
compose-002 & H & Check SF and NYC weather; email Alice if $<$40°F with March~1 events. & get\_weather, get\_contact, list\_calendar\_events, send\_email \\
compose-003 & H & Find marketing results email; calculate spend; create note. & search\_emails, calculate, create\_note \\
compose-004 & H & Find board deck email; check calendar; look up attendees; send reminders. & search\_emails, list\_calendar\_events, get\_contact, send\_email \\
\midrule
\multicolumn{4}{@{}l}{\textit{Ambiguous}} \\
ambig-001 & M & Help me prepare for my meetings tomorrow. & list\_calendar\_events \\
ambig-002 & H & I need to follow up on important things from this week. & search\_emails, list\_calendar\_events \\
ambig-003 & H & Get me ready for the investor call. & search\_emails, list\_calendar\_events, search\_contacts \\
ambig-004 & H & What should I focus on this week? & list\_calendar\_events, search\_emails \\
\end{longtable}
\normalsize

\clearpage
\section*{Appendix B. Tool Descriptions}
\addcontentsline{toc}{section}{Appendix B. Tool Descriptions}
\label{app:tools}

Table~\ref{tab:tools} summarizes the 10 deterministic simulated tools used in
our environment. Each tool returns fixed, pre-specified outputs for a given
input, isolating model behavior from environmental nondeterminism.

\small
\begin{longtable}{@{}llp{0.52\textwidth}@{}}
\caption{Deterministic simulated tools used in the benchmark.}\label{tab:tools}\\
\toprule
Tool & Domain & Description \\
\midrule
\endfirsthead
\toprule
Tool & Domain & Description \\
\midrule
\endhead
\midrule
\multicolumn{3}{r}{\textit{Continued on next page}} \\
\endfoot
\bottomrule
\endlastfoot
\texttt{get\_contact}            & Contacts & Look up a contact by name; returns email, phone, and company. \\
\texttt{search\_contacts}        & Contacts & Search contacts by free-text query; returns a matching list. \\
\texttt{send\_email}             & Email    & Send an email with fields such as recipient, subject, and body; returns a confirmation. \\
\texttt{search\_emails}          & Email    & Search the inbox by query; returns matching message records. \\
\texttt{list\_calendar\_events}  & Calendar & List events for a date or date range. \\
\texttt{create\_calendar\_event} & Calendar & Create an event with title, date, time, duration, and attendees. \\
\texttt{search\_products}        & Products & Search inventory; returns price, stock, and category metadata. \\
\texttt{calculate}               & Math     & Evaluate a mathematical expression and return the numeric result. \\
\texttt{get\_weather}            & Weather  & Return current weather for a city, including temperature and conditions. \\
\texttt{create\_note}            & Notes    & Create a text note with title and body; returns a confirmation. \\
\end{longtable}
\normalsize

\clearpage
\section*{Appendix C. System Prompt}
\addcontentsline{toc}{section}{Appendix C. System Prompt}
\label{app:system_prompt}

The following system prompt was used verbatim for all models and all runs:

\begin{quote}
\small\ttfamily
You are a helpful assistant with access to a set of tools. Use the provided tools to complete the user's request as accurately and efficiently as possible. Only call tools when necessary. If you have all the information you need to respond, do so directly. When you have completed the task, provide a concise summary of what you did and the results.
\end{quote}

This prompt is intentionally minimal so that we do not anchor models to a
particular tool-selection strategy and thereby artificially inflate measured
consistency.

\clearpage
\section*{Appendix D. Correctness Criteria (Illustrative Subset)}
\addcontentsline{toc}{section}{Appendix D. Correctness Criteria (Illustrative Subset)}
\label{app:correctness}

Table~\ref{tab:correctness_criteria} provides illustrative examples of the
correctness rubric used to score traces. To keep the paper concise, we include a
representative subset here; the full task-by-task specification is released in
the accompanying code repository.

\small
\begin{longtable}{@{}lp{0.28\textwidth}p{0.42\textwidth}@{}}
\caption{Illustrative correctness criteria for representative tasks.}\label{tab:correctness_criteria}\\
\toprule
Task ID & Required Tools & Key Argument Checks \\
\midrule
\endfirsthead
\toprule
Task ID & Required Tools & Key Argument Checks \\
\midrule
\endhead
\midrule
\multicolumn{3}{r}{\textit{Continued on next page}} \\
\endfoot
\bottomrule
\endlastfoot
retrieve-001 & get\_contact, send\_email & \texttt{get\_contact.name}$\sim$\texttt{/alice/i}; \texttt{send\_email.to}=\texttt{alice@...}; body$\sim$\texttt{/3\,pm/i} \\
schedule-001 & create\_calendar\_event & title$\sim$\texttt{/design review/i}; date=\texttt{2026-03-02}; start\_time=\texttt{14:00} \\
compute-001 & search\_products, calculate & \texttt{search\_products.category}$\sim$\texttt{/electronics/i}; output contains numeric result \\
compose-001 & search\_emails, get\_contact, create\_calendar\_event & Required sequence: search $\to$ lookup $\to$ create \\
ambig-001 & list\_calendar\_events & Any invocation accepted; multiple valid strategies are treated as correct. \\
\end{longtable}
\normalsize

\end{document}